\documentclass[conference]{IEEEtran}
\IEEEoverridecommandlockouts
\usepackage{cite}
\usepackage{amsmath,amssymb,amsfonts}
\usepackage{algorithmic}
\usepackage{graphicx}
\usepackage{textcomp}
\usepackage{xcolor}
\usepackage{float}
\usepackage{gensymb}
\usepackage{siunitx}
\usepackage{fancyhdr}

\def\BibTeX{{\rm B\kern-.05em{\sc i\kern-.025em b}\kern-.08em
    T\kern-.1667em\lower.7ex\hbox{E}\kern-.125emX}}

\makeatletter

\def\ps@IEEEtitlepagestyle{%
  \def\@oddfoot{\mycopyrightnotice}%
  \def\@evenfoot{}%
}
\def\mycopyrightnotice{%
  {\footnotesize 978-1-6654-6962-3/22/\$31.00 \textcopyright2022 IEEE \hfill}
  \gdef\mycopyrightnotice{}
}

\begin{document}

\title{Face Recognition In Children: A Longitudinal Study\\

\thanks{This material is based upon work supported by the Center for Identification Technology Research and the National Science Foundation (NSF) under Grant No.$650503$. Authors would also like to thank the \textit{Chameleon} cloud for providing the computational infrastructure required for this work \cite{keahey_lessons_2020}. 
}
}

\author{\IEEEauthorblockN{Keivan Bahmani, Stephanie Schuckers}
\IEEEauthorblockA{\textit{Clarkson University} \\
Potsdam, NY, USA \\
bahmank@clarkson.edu, sschucke@clarkson.edu}
}

\maketitle

\IEEEpubidadjcol

\thispagestyle{fancy}
\fancyhf{}
\lhead{\textcolor{red}{To appear in 10th IEEE International Workshop on Biometrics and Forensics - IWBF 2022 - Salzburg, Austria, April 20-21, 2022}}  

\begin{abstract}
The lack of high fidelity and publicly available longitudinal children face datasets is one of the main limiting factors in the development of face recognition systems for children. In this work, we introduce the Young Face Aging (YFA) dataset for analyzing the performance of face recognition systems over short age-gaps in children. We expand previous work by comparing YFA with several publicly available cross-age adult datasets to quantify the effects of short age-gap in adults and children. Our analysis confirms a statistically significant and matcher independent decaying relationship between the match scores of ArcFace-Focal, MagFace, and Facenet matchers and the age-gap between the gallery and probe images in children, even at the short age-gap of 6 months. However, our result indicates that the low verification performance reported in previous work might be due to the intra-class structure of the matcher and the lower quality of the samples. Our experiment using YFA and a state-of-the-art, quality-aware face matcher (MagFace) indicates 98.3\% and 94.9\% TAR at 0.1\% FAR over 6 and 36 Months age-gaps, respectively, suggesting that face recognition may be feasible for children for age-gaps of up to three years.


\end{abstract}


%
\IEEEpeerreviewmaketitle

\section{Introduction}
Aging is a natural part of every human's life. However, this natural and unavoidable process is an obstacle in performing automatic face recognition across ages. For most face recognition applications, it is expected that the time between enrollment and matching may be many months or years.
In Face Recognition (FR) systems, the within-identity variation can be due to subjects' pose, illumination, expressions and aging \cite{lanitis_survey_2010}. Previous work on FR systems suggest that aging is one of the most significant sources of within-identity variation \cite{lanitis_survey_2010, otto_how_2012, ricanek_morph_2006}. Additionally, unlike other sources of within-identity variation such as lighting or pose, aging is inevitable and cannot be controlled during the image acquisition phase. The current state of the art FR systems rely on deep convolutional neural network (CNNs)-based FR systems perform very well across different poses, illumination, and expressions i.e., \textit{in the wild environment} \cite{huang_labeled_2014, grother_face_2019}. 
However, such systems are still susceptible to issues related to aging and suffer more than 10\% drop in accuracy when evaluated at large age gaps \cite{zheng_cross-age_2017}. The recent NIST Face Recognition Vendor Test (FRVT) also identified age-related degradation in the performance of all the evaluated algorithms over large age gaps in adults \cite{grother_face_2019}. Currently, several publicly available and longitudinal datasets are available to researchers to facilitate the development of cross-age face recognition systems in adults \cite{chen_face_2015, zheng_cross-age_2017, moschoglou_agedb_2017, ricanek_morph_2006, bianco_large_2017}. However, aging in children is physiologically different. The preponderance of aging in adults is due to soft tissue changes i.e. skin texture, wrinkles \cite{ricanek_jr_craniofacial_2008}. On the other hand, aging in children involves a non-linear cranial growth, i.e changes in the bone structure of the skull that changes the appearance of the face \cite{ramanathan_modeling_2006, ricanek_review_2015}. As a result, face recognition in children requires special considerations, and models developed on adults may not always be applicable to children.

\begin{figure}[t]
\begin{minipage}[b]{0.24\linewidth}
    \begin{center}
        \includegraphics[width=1\linewidth]{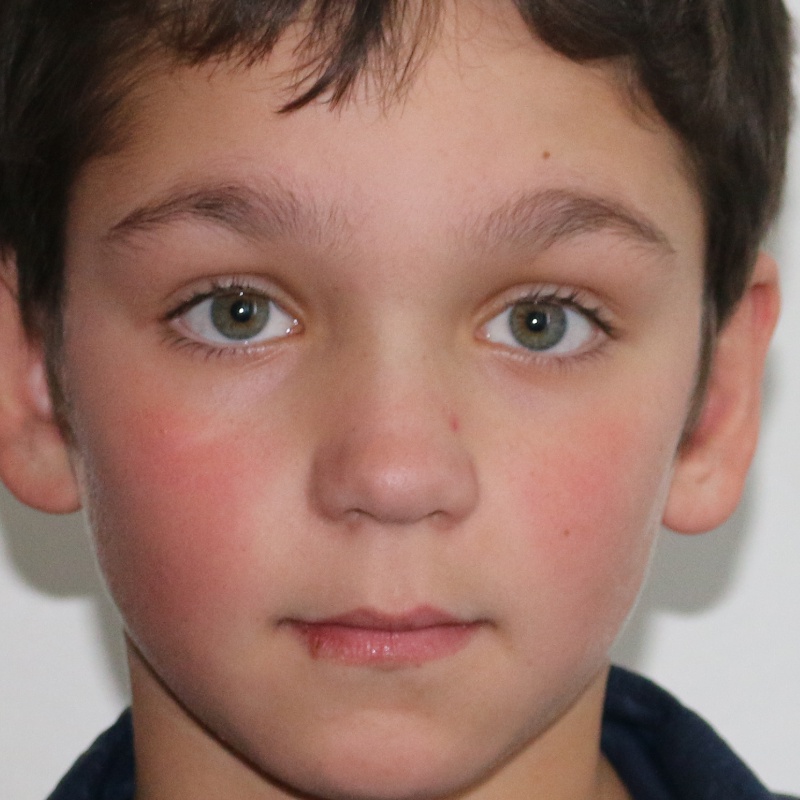}
    \end{center}
  \centerline{(a)}
\end{minipage}
\begin{minipage}[b]{0.24\linewidth}
    \begin{center}
        \includegraphics[width=1\linewidth]{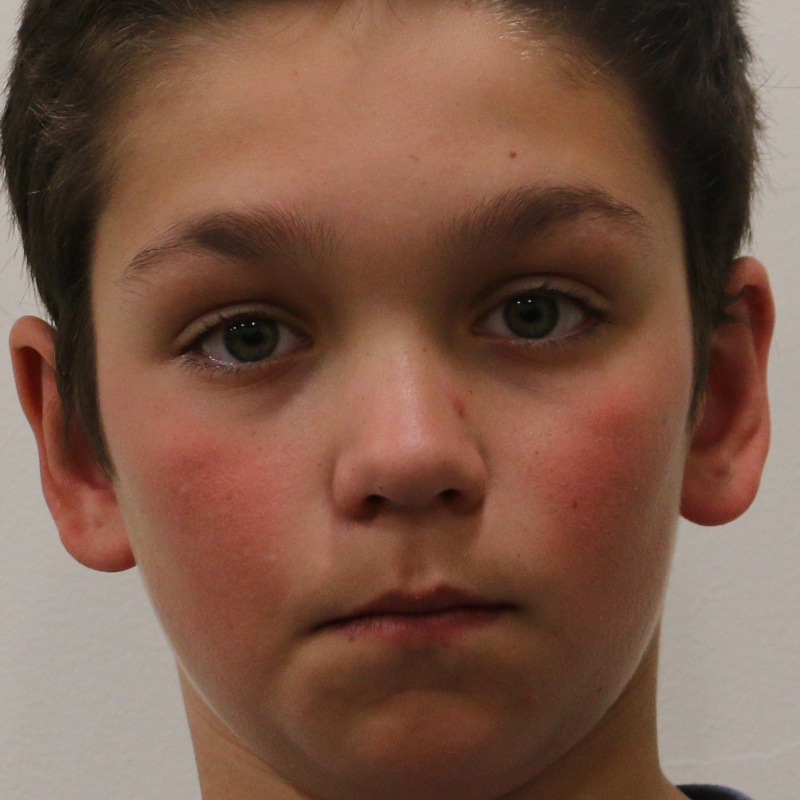}
    \end{center}
  \centerline{(b)}
\end{minipage}
\begin{minipage}[b]{0.24\linewidth}
    \begin{center}
        \includegraphics[width=1\linewidth]{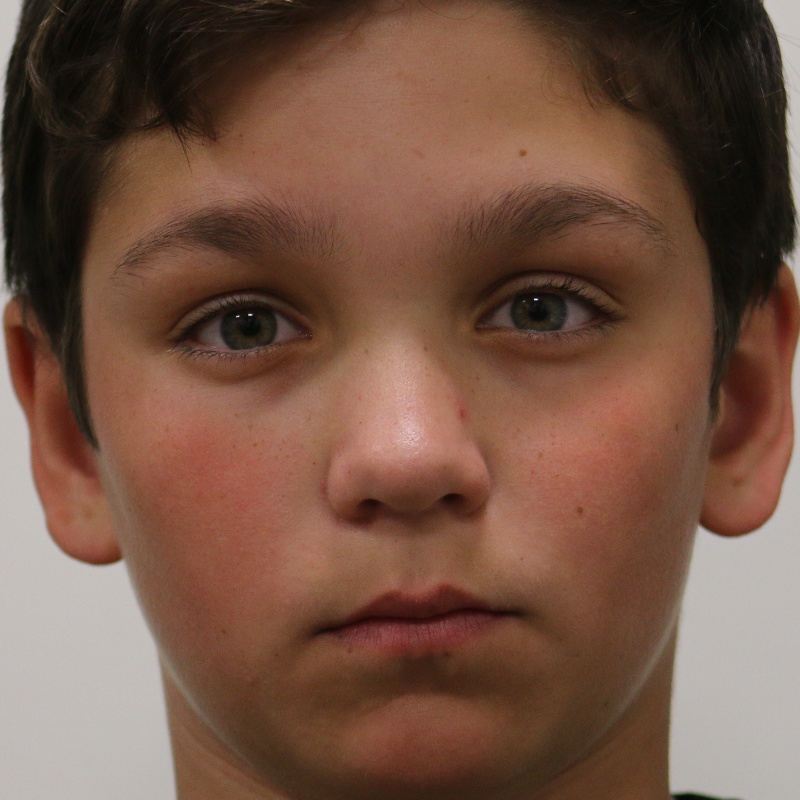}
    \end{center}
  \centerline{(c)}
\end{minipage}
\begin{minipage}[b]{0.24\linewidth}
    \begin{center}
        \includegraphics[width=1\linewidth]{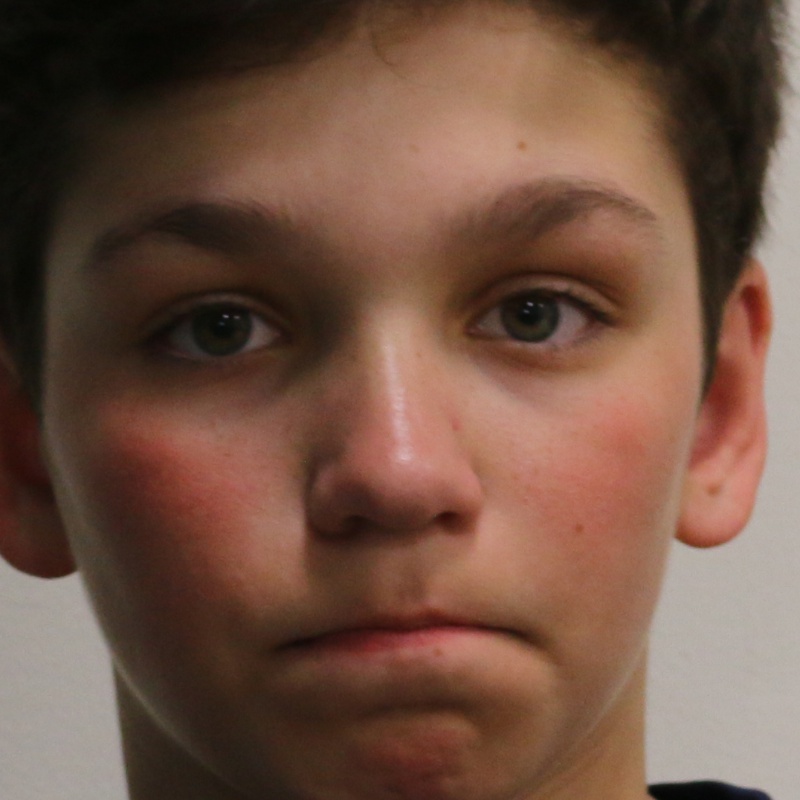}
    \end{center}
  \centerline{(d)}
\end{minipage}
\caption{Age progression of a subject in YFA: a) template b) 12 months, c) 24 months, d) 36 months. \textbf{Best viewed in color}}
\label{fig:YFA_Samples1}
\end{figure}
\begin{figure}[t]
\begin{minipage}[b]{0.24\linewidth}
    \begin{center}
        \includegraphics[width=1\linewidth]{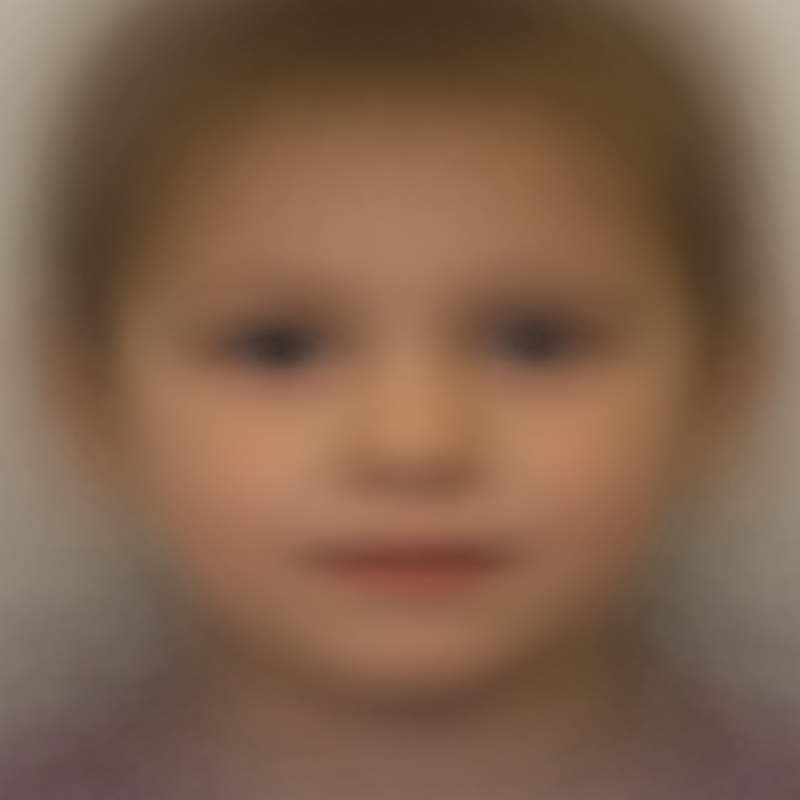}
    \end{center}
  \centerline{(a)}
\end{minipage}
\begin{minipage}[b]{0.24\linewidth}
    \begin{center}
        \includegraphics[width=1\linewidth]{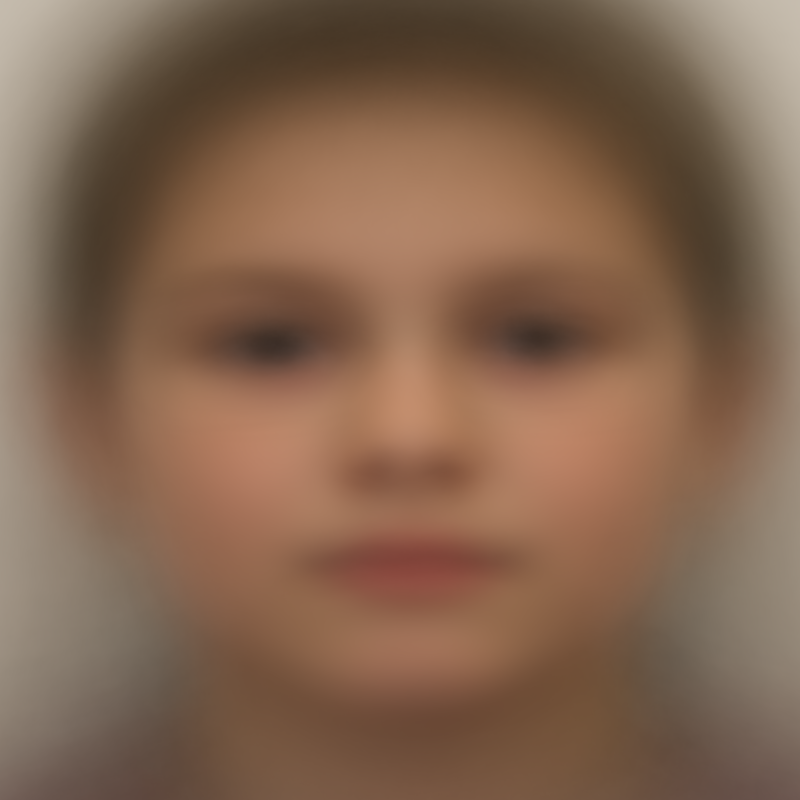}
    \end{center}
  \centerline{(b)}
\end{minipage}
\begin{minipage}[b]{0.24\linewidth}
    \begin{center}
        \includegraphics[width=1\linewidth]{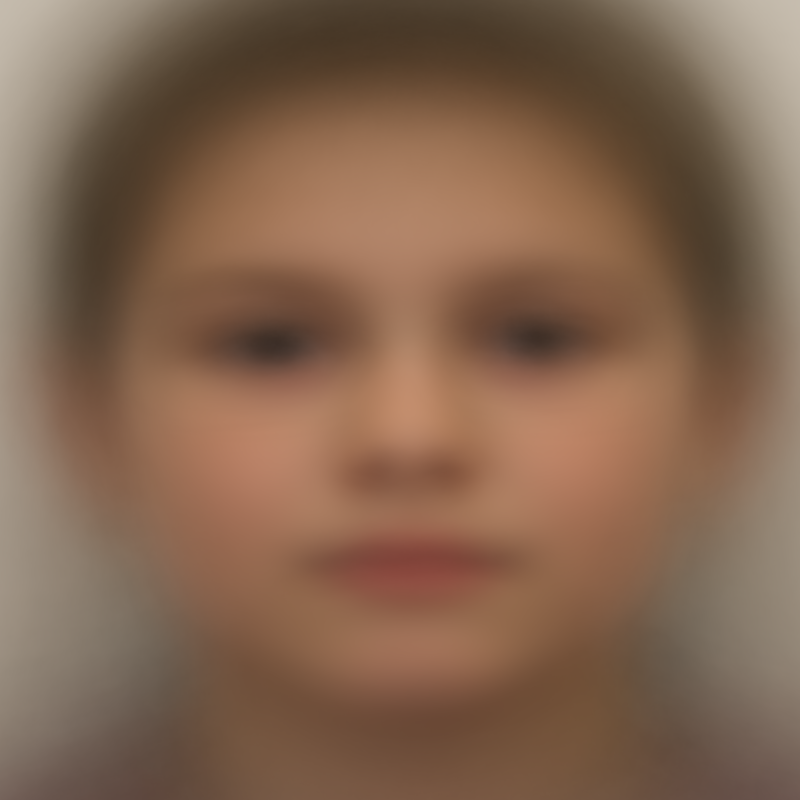}
    \end{center}
  \centerline{(c)}
\end{minipage}
\begin{minipage}[b]{0.24\linewidth}
    \begin{center}
        \includegraphics[width=1\linewidth]{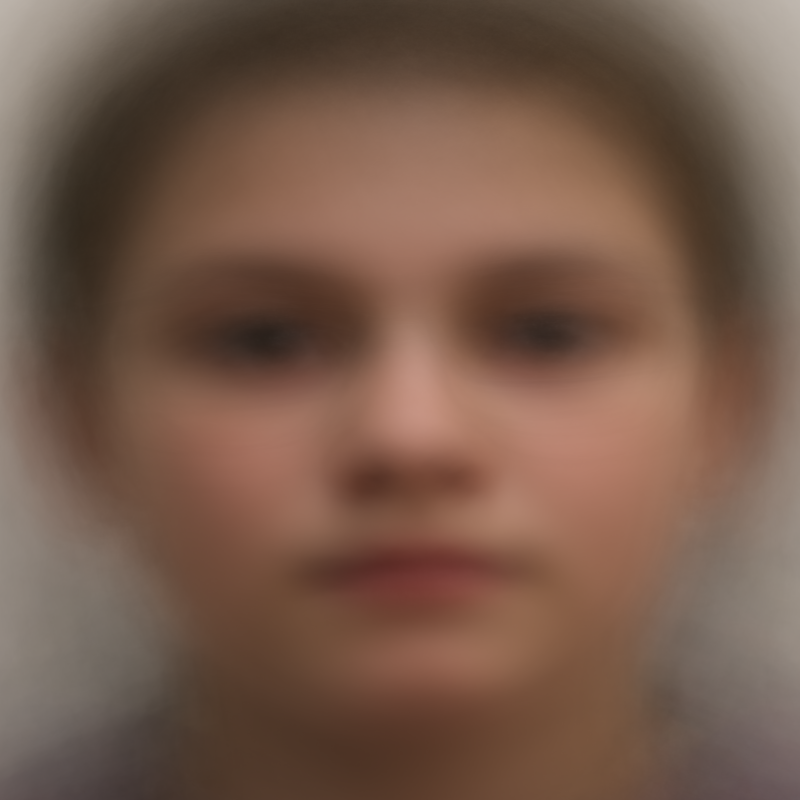}
    \end{center}
  \centerline{(d)}
\end{minipage}
\caption{Age progression in the average of YFA samples:  a) Avg. of samples younger than 6 years old b) Avg. of 6 to 8 years old samples , c) Avg. of 9 to 11 years old samples d) Avg. of 12 to 14 years old samples. \textbf{Best viewed in color}}
\label{fig:YFA_Samples2}
\end{figure}

\begin{table*}[ht]
\begin{center}
\begin{tabular}{|p{1.9cm}|| p{1.5cm}| p{1.5cm}| p{2.2cm}| p{1.7cm} |p{2cm}| p{1.9cm}|} 
\hline
Datasets & \# Subjects & \# Samples & \# Images/Subject & Age (Years) & Age Gap & Environment\\
\hline
\hline
CACD-VS \cite{chen_face_2015} &  4,000 Pairs & 8,000  & N.A. & 10 - 60+ & 0 – 10 & In the Wild \\
\hline
AgeDB  \cite{moschoglou_agedb_2017} & 568 & 16,488 & 29.0 & 1 - 101 & 0 - 90 &  In the Wild\\
\hline
MORPH-II \cite{ricanek_morph_2006} & 13,000  & 55,133  & 4.1 & 16 - 77 & 0 - 5  & Controlled \\
\hline
LAG \cite{bianco_large_2017} &   1,010 & 3,828 & 3.8 & N.A. & Child to Adult &  In the Wild\\
\hline
YFA (Ours) &  231 &  2293 & 9.9 & 3 - 14 & 0 - 3 & Controlled\\
\hline
\end{tabular}
\end{center}
\label{table:Datasets}
\caption{Summary of the cross-age face recognition datasets used in this work.}
\end{table*}

Several previous work aimed to quantify the effects of aging in state-of-the-art children face recognition systems. Ricanek et al. developed In-The-Wild Child Celebrity (IWCC) - a dataset of 1705 images from 304 subjects from 5 to 32 years old. Authors investigated the performance of several hand-crafted FR algorithms and reported a low performance of 37\% True Accept Rate (TAR) at 1\% False Accept Rate (FAR) \cite{ricanek_review_2015}. Ramanathan et al. showed that transforming the images using a non-linear craniofacial growth model can improve the performance of \textit{eigenface} model in the \textit{FGNET} dataset \cite{ramanathan_modeling_2006}. Srinivas et al identified the differential performance in several COTS and open-source face recognition models between adults and children using in the wild datasets, where the best performing model achieved 0.782 TAR at 0.01\% FAR \cite{srinivas_face_2019}. However, due to the use of COTS matchers, this work does not provide any details on the underlying algorithms or factors behind the low performance in children. Deb et al. introduced Children Longitudinal Face (CLF) dataset. CLF contains 3682 images from 919 West Asian children between 2 to 18 years old, with minimum age-gap of 2 years and maximum age gap of 7 Years. The dataset is captured at 354×472 pixels and has challenging variations in terms of pose, illumination, expression, and obstructions \cite{deb2018longitudinal}. Authors evaluated the verification performance of an open source version of the Facenet matcher \cite{schroff_facenet:_2015} trained on MS-Celeb \cite{guo2016ms} and reported a 43.87\% TAR at 0.01\% FAR. After fine tuning the Facenet matcher with another private children dataset, the verification performance improved to 57.7\% TAR at 0.01\% FAR. Additionally, authors developed a statistical model for the changes in the genuine comparison scores based on covariates such as age-gap, enrolment age, and gender of the subjects. To the best of our knowledge, \cite{michalski_impact_2018} is the only previous work that is carried out using an operational high quality, and controlled dataset. Authors evaluated the performance of the \textit{NeoFace 3.1} commercial face recognition algorithm over a large and controlled dataset with more than 3M samples from children under 17 years old. However, their experiment using a fixed threshold based on 0.1\%
False Match Rate (FMR) operational point in adults, respectively lead to FAR of 1.3\%, 1\%, and 0.7\% at 91.8\%, 93.1\% and 95.2\% TAR for four, five and six years old children at age-gap of three years. Given the relatively higher performance observed in a single COTS matcher using high quality data \cite{michalski_impact_2018} and lower performance reported using uncontrolled in-the-wild data (lower quality) \cite{ramanathan_modeling_2006, ricanek_review_2015, deb2018longitudinal, srinivas_face_2019}, it is difficult to disentangle the effects of aging from other within-identity factors affecting the performance of face matchers. Additionally, it is not clear how recent advanced face matchers designed to curtail the within-identity factors \cite{deng_arcface_2019, wang2021face, meng_magface_2021} would perform using high quality data and what factors affect their performance. This is important as many applications, e.g., benefit distribution, border security, etc., rely on high quality photos such as passports.

In this work, we present the Young Face Aging (YFA) dataset. YFA samples are captured under consistent indoor lighting, expression, and pose. Figures \ref{fig:YFA_Samples1} and \ref{fig:YFA_Samples2} respectively depict the age progression of an individual and the average of subjects over YFA. The controlled image acquisition environment of the YFA allows us to better disentangle the effects of aging from other within-identity variations. Additionally, YFA extends the previous work by capturing the subjects at a higher frequency (every 6 months). We utilized YFA in conjunction with CACD-VS\cite{chen_face_2015}, AgeDB\cite{moschoglou_agedb_2017}, LAG\cite{bianco_large_2017}, and Morph-II \cite{ricanek_morph_2006} cross-age adult datasets to provide a comprehensive analysis of the biometric performance of Facenet, VGGFace, VGGFace2, ArcFace, ArcFace-Focal, and MagFace face recognition models in both adults and children.

We evaluated the verification performance of Facenet \cite{schroff_facenet:_2015}, VGGFace\cite{parkhi_deep_2015}, VGGFace2\cite{cao2018vggface2}, ArcFace\cite{deng_arcface_2019}, ArcFace-Focal \cite{wang2021face}, and MagFace\cite{meng_magface_2021} models using the YFA dataset. Our result confirms that degradation of performance observed in previous work using the older models even in a short age-gap of 6 month. However, contrary to previous work using in-the-wild datasets, our result indicates that a combination of a quality-aware face recognition model such as Magface and high-quality samples (YFA) can result in a high TAR of 98.3\% (after one year) and 94.9\% (after 3 years) at FAR of 0.1\%. 

Additionally, we evaluated the impact of age-gap, enrolment age, and gender using Linear Mixed Effect (LME) modeling. Our analysis respectively indicates an estimated decrease in the MS, due to age-gap of $-0.033\pm0.003$ (approx), $-0.042\pm0.003$ (approx), and $-0.028 \pm 0.002$ (approx) per year for Magface, Arcface-Focal, and Facenet-V1 face recognition models. Our analysis did not find consistent and matcher independent relationships between match score and enrolment age or gender in YFA.

Finally, we utilize the \textit{BEAT} platform \cite{anjos_beat_2017} to make the YFA the first publicly available longitudinal, high resolution and controlled children face aging dataset. The \textit{BEAT} platform allows the research community to utilize the YFA while preserving the privacy of the subjects.

\begin{table*}[!ht]
\begin{center}
\begin{tabular}{|p{2.2cm}|p{1.9cm}|p{1.3cm}|p{1.1cm}|p{1.3cm}|p{1.4cm}| p{1cm}| p{1cm}| p{1cm}|p{1cm}} 
\hline
\multicolumn{4}{|c|}{Face Matchers} & \multicolumn{5}{|c|}{Datasets}\\
\hline
Matcher & Training Dataset & Input size & \#Features& CACD-VS & MORPH-II & AgeDB & LAG* & \textbf{YFA}\\
\hline
\hline
Facenet-V1 \cite{schroff_facenet:_2015} &  MS-Celeb \cite{guo2016ms} & $160 \times 160$ & 128 & 0.944  & 0.970 & 0.484 & 0.105 & 0.708\\
\hline
Facenet-V2 \cite{schroff_facenet:_2015} &  VGG-Face2 \cite{cao2018vggface2} & $160 \times 160$ & 512 & 0.586 & 0.708 & --- & --- &    0.293\\
\hline
VGGFace \cite{parkhi_deep_2015} &  LFW \cite{huang_labeled_2014} & $224 \times 224$ & 4096 & 0.396 & 0.215 & --- & ---  & 0.312\\
\hline
VGGFace2 \cite{cao2018vggface2} &  VGG-Face2 \cite{cao2018vggface2} & $224 \times 224$ & 512 & 0.549 & 0.526 & 0.195 & 0.081 &0.331\\
\hline
ArcFace \cite{deng_arcface_2019} &  MS1MV2 \cite{deng_arcface_2019} & $112 \times 112$ & 512 & 0.770 & 0.968 & 0.379 & 0.231  &  0.772\\
\hline
ArcFace-Focal \cite{wang2021face} &  MS1MV2 \cite{deng_arcface_2019} & $112 \times 112$ & 512 & \textbf{0.964} & \textbf{0.989} & \textbf{0.681} & \textbf{0.297} & 0.854\\
\hline
MagFace \cite{meng_magface_2021} &  MS1MV2 \cite{deng_arcface_2019} & $112 \times 112$ & 512 & 0.832 & 0.968 & 0.402 & 0.242 & \textbf{0.933}\\
\hline
\end{tabular}
\end{center}
\label{table:Matchers}
\caption{Face Matcher description including training dataset, input size, and \# features (Left four columns). Verification performance (TAR at 0.01\% FAR) (Right five columns). "---" indicates that the matcher did not achieve 0.01\% FAR. "*" The verification of performance for the LAG dataset is calculated using only young to adult genuine comparisons.}
\end{table*}

\section{METHODOLOGY}
We believe explaining the effects of short age-gap in children is more meaningful in comparison to other datasets that include just adults or adults/children.
Consequently, we utilized five cross-age adult datasets with different age gaps and the same pre-processing as YFA to adequately present the performance of the evaluated face matchers. This allows us to compare 
the impact of aging in children face recognition with respect to short/long aging in adults. Subsequently, we use Linear Mixed-Effect (LME) models \cite{lindstrom_newtonraphson_1988} to investigate the influence of enrolment age, age-gap, and gender on the match score of children FR systems. The rest of this section provides a brief overview of the datasets and face matchers used in this work.

\subsection{Cross-Age Datasets}
In this section, we present a brief overview of the cross-age datasets used in this work. Table \ref{table:Datasets} summarizes the attributes of the datasets.

\subsubsection{CACD\_VS} The Verification Section of the Cross-Age Celebrity Dataset (CACD-VS) contains 2000 mated, 2000 non-mated pairs for 2000 celebrities. CACD\_VS is annotated by human annotators to confirm the correct identity \cite{chen_face_2015}.
 
\subsubsection{MORPH-II}
The  academic  MORPH  dataset is a non-commercial longitudinal and controlled dataset collected over 5 years (2003 - 2007). The average number of images per individual is 4.1 with an average of 164 days between the captures \cite{ricanek_morph_2006}. This dataset is used to evaluate the performance of short age-gap (up to 5 years) in adults.

\subsubsection{AgeDB} AgeDB is in-the-wild cross-age dataset with large maximum age gap of 90 years. Samples in AgeDB  are manually collected by humans as opposed to other automatically gathered datasets. This dataset provides 183 samples from 109 children under 16. As a result, AgeDB can provide us with both adult and children samples \cite{moschoglou_agedb_2017}.

\begin{table*}[!ht]
\begin{center}
\begin{tabular}{|p{1.7cm}|| p{1cm} | p{1.2cm} |p{1.3cm} | p{1.3cm}| p{1.3cm}| p{1.3cm}| p{1.3cm}| p{1.3cm}|} 
\hline
Model & FAR & Threshold & $\Delta T$ = 6M & $\Delta T$ = 12M & $\Delta T$ = 18M & $\Delta T$ = 24M & $\Delta T$ = 30M & $\Delta T$ = 36M \\
\hline
\hline
Facenet-V1 & 0.1\% & 0.630  & 95.8 & 94.8  &  92.5 & 84.3 & 82.7 & 76.0 \\
\hline
Facenet-V1 & 0.01\% & 0.826 & 85.6 & 80.1  & 74.5  & 64.8 & 57.0 & 43.4 \\
\hline
\hline
ArcFace & 0.1\% & 0.474& 87.6  & 88.1  & 85.3  & 84.8 & 86.3 & 81.1 \\
\hline
ArcFace & 0.01\% & 0.556 & 81.0 &  78.8 & 78.1  & 74.8 & 75.6 & 69.9 \\
\hline
\hline
ArcFace-Focal & 0.1\% & 0.532 & 97.6  & 98.3  & 95.4  & 92.7 & 93.1 & 91.6 \\
\hline
ArcFace-Focal & 0.01\% & 0.630 & 92.8 &  91.5 & 87.9  & 79.5 & 78.5 & 72.8 \\
\hline
\hline
MagFace & 0.1\% & 0.453 & \textbf{98.2} & \textbf{98.3}  & \textbf{98.0}  & \textbf{97.2}  & \textbf{97.3} & \textbf{94.9} \\
\hline
MagFace & 0.01\% & 0.549 & \textbf{96.9} & \textbf{95.2} & \textbf{93.2} & \textbf{91.6} & \textbf{92.9} & \textbf{84.7} \\
\hline
\end{tabular}
\end{center}
\label{table:ComparisonTPR}
\caption{Verification performance (TAR at 0.01\% and 0.1\% FAR) of Facenet-V1, ArcFace, ArcFace-Focal, and MagFace models for increasing age-gap (6-36 Months) between enrollment and query samples in YFA dataset.}
\end{table*}

\begin{table*}[ht]
\begin{center}
\begin{tabular}{|p{1.2cm}|| p{1.5cm}| p{3.2cm}||p{3.2cm}|| p{3.2cm}|}
\hline
\multicolumn{5}{|c|}{\textbf{Morph-II}}\\
\hline
\hline
Variable & Parameter & MagFace $(Est\pm SE)$ & ArcFace-Focal $(Est\pm SE)$ & Facenet-V1 $(Est\pm SE)$\\
\hline
\hline
Intercept & $\beta_0$ & $0.700 \pm 0.002^{***}$  & $0.796 \pm 0.002^{***}$ & $0.862 \pm 0.001^{***}$ \\
\hline
$\Delta T$ & $\beta_1$ & $-0.11 \pm 0.000^{***}$ & $-0.010 \pm 0.000^{***}$ & $-0.008 \pm 0.000^{***}$\\
\hline
EA & $\beta_2$ & $0.000 \pm 0.000^{***}$ & $0.000 \pm 0.000 NS$ & $-0.000 \pm -0.000*$\\
\hline
G & $\beta_3$ & $-0.010 \pm 0.002^{***}$ & $-0.038 \pm 0.002^{***}$ & $-0.025 \pm 0.002^{***}$\\
\hline
\hline
\multicolumn{5}{|c|}{\textbf{AgeDB - Adults}} \\
\hline 
\hline 
Intercept & $\beta_0$ & $0.374 \pm 0.004^{***}$ & $0.512 \pm 0.006^{***}$ & $0.649 \pm 0.002^{***}$ \\
\hline
$\Delta T$ & $\beta_1$ & $-0.004 \pm 0.000^{***}$ & $-0.004 \pm  0.000^{***}$ & $-0.005 \pm 0.001^{***}$ \\
\hline
EA & $\beta_2$ & $0.001 \pm 0.000^{***}$ & $0.001 \pm 0.000^{***}$  & $0.001 \pm 0.000^{***}$ \\
\hline
G & $\beta_3$ & $-0.007 \pm 0.004 NS$ & $-0.005 \pm 0.003 NS$ &  $-0.004 \pm 0.007 NS$ \\
\hline
\hline
\multicolumn{5}{|c|}{\textbf{AgeDB - Young}}\\
\hline
\hline 
Intercept & $\beta_0$ & $0.439 \pm  0.045 ^{***}$ & $0.412 \pm 0.053^{***}$ &  $0.698 \pm  0.042^{***}$ \\
\hline
$\Delta T$ & $\beta_1$ & $-0.022 \pm 0.004 ^{***}$ & $-0.024 \pm 0.005^{***}$ &  $-0.026 \pm  0.004^{***}$\\
\hline
EA & $\beta_2$ & $-0.000 \pm 0.002 NS$ & $0.008 \pm 0.002^{**}$ & $0.001 \pm 0.002 NS$ \\
\hline
G & $\beta_3$ & $-0.055 \pm 0.029 NS$ & $-0.077 \pm 0.037^{*}$  &  $-0.075 \pm  0.031^{*}$ \\
\hline
\hline
\multicolumn{5}{|c|}{\textbf{Young Face Aging (YFA)}}\\
\hline
\hline
Intercept & $\beta_0$ & $0.745 \pm 0.018^{***}$ & $0.757 \pm 0.018^{***}$ & $0.897 \pm 0.012^{***}$\\
\hline
$\Delta T$  & $\beta_1$ &$ -0.033 \pm 0.003^{***}$ & $-0.042 \pm 0.003^{***}$ & $-0.028 \pm 0.002^{***}$ \\
\hline
EA & $\beta_2$ & $0.000 \pm 0.002 NS$ & $0.006 \pm 0.002^{**}$ & $-0.001 \pm 0.000  NS$ \\
\hline 
G & $\beta_3$ & $0.003 \pm 0.010 NS$ & $-0.010 \pm 0.011 NS$ & $-0.002 \pm 0.007 NS$ \\
\hline
\end{tabular}
\end{center}
\label{table:YFA}
\caption{Fixed effects of the  Facenet-V1, ArcFace-Focal, and MagFace models evaluated on the Morph-II, Age-DB, and YFA Datasets. Significance Code: 0 ‘***’ 0.001 ‘**’ 0.01 ‘*’ 0.05 ‘.’ 0.1 ‘ ’ 1 ; *** indicates p-value between 0 and 0.001 with significance level 0.001 and so on.
Est.: Estimate, SE: Standard Error, NS: Not significant}
\end{table*}

\subsubsection{Large Age Gap (LAG)} This dataset does not provide accurate age labels. However, each subject has at least one sample denoted by authors as "child/young" and multiple samples denoted as adults. As a result, LAG can be used to represent the performance of matchers in a very challenging task of large age-gap face recognition between children and adults.

\subsection{Young Face Aging (YFA) dataset}
\begin{figure}[!h]
\begin{minipage}[b]{0.48\linewidth}
    \begin{center}
        \includegraphics[width=1\linewidth]{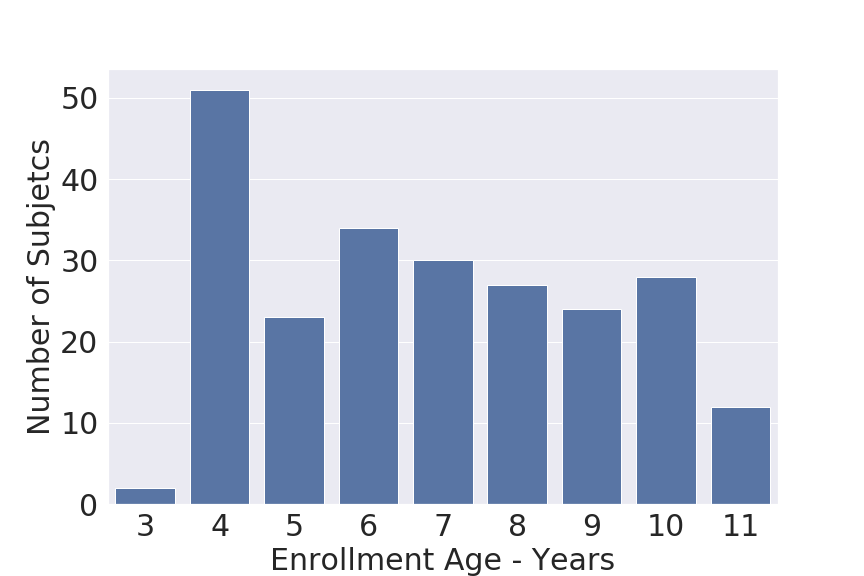}
    \end{center}
\end{minipage}
\begin{minipage}[b]{0.48\linewidth}
    \begin{center}
        \includegraphics[width=1\linewidth]{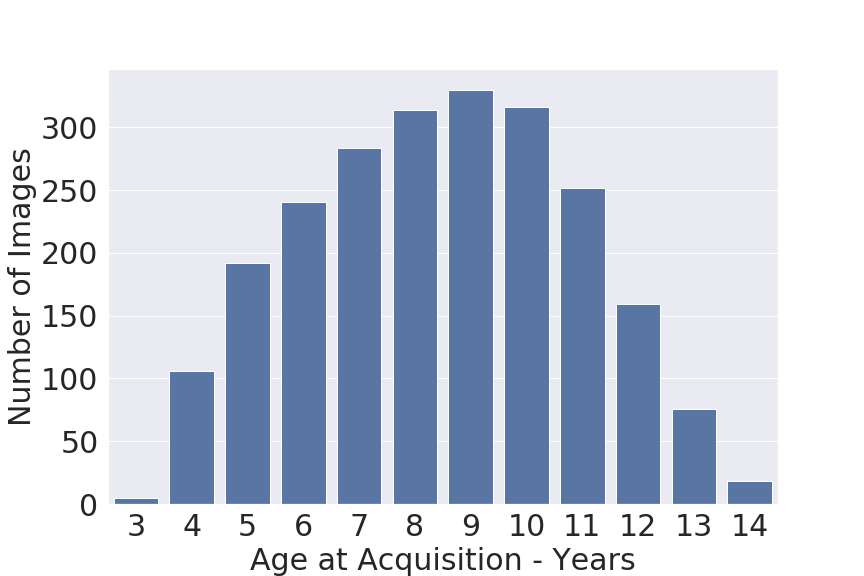}
    \end{center}
\end{minipage}
\caption{Statistics of the YFA Dataset. Right: age at acquisition (years). Left: age at the enrolment (years)}
\label{fig:YFA_stats}
\end{figure}

Young Face Aging (YFA) dataset contains 2293 samples from 231 subjects collected in a controlled environment with a time-lapse of 6 months over the period of 3 years. Figure \ref{fig:YFA_stats} depicts the statistics of the YFA. Samples are captured from 3-14 years old children. The research team collaborates with the local elementary and middle school to identify and enroll subjects for voluntary participation, in accordance with an approved IRB protocol. Images are captured using a DSLR camera at the resolution of 3648 by 5472 pixels. The image acquisition is carried out with consistent indoor lighting, natural expression, and minimal variation in the subject's pose. Each subject is captured at least twice during each session. The Year of birth (or grade if the year is unavailable) for each subject is recorded during the enrolment process and has been used to record the age at subsequent acquisitions. YFA has a maximum age-gap of 3 years with 135 out of 231 subjects (58\%) reaching the maximum age gap in the dataset. The average number of samples per subject is 9.92. Most subjects are of Caucasian background; however, YFA does not provide self-reported ethnicity labels. YFA is balanced in terms of gender (117 Female, 114 Male), 2096 samples have no glasses (91.4\%), while 197 samples have glasses (8.6\%). 

\subsection{Face Detection and Recognition Models}
We utilize the same face detection and processing pipeline on all the evaluated datasets to reduce the effects of the pre-processing on the performance of the evaluated matchers. All the images are processed using the MTCNN face detection model \cite{zhang_joint_2016} to detect and align the face. We enforce a very tight crop around the face to minimize the effects of background noise.
Subsequently each cropped face has been resized to the input requirements of each specific matcher. We evaluated the verification performance of each dataset using two versions of the open source implementation of the Facenet matcher\footnote{$https://github.com/davidsandberg/facenet$} (Facenet-V1 is trained using centerloss while Facenet-V2 is trained using softmax loss \cite{wen2019comprehensive}),  VGGFace\footnote{$https://www.robots.ox.ac.uk/\sim vgg/software/vgg\_face/$} 
\cite{parkhi_deep_2015}, VGGFace2\footnote{$https://github.com/WeidiXie/Keras-VGGFace2-ResNet50$} \cite{cao2018vggface2}, 
Arcface\footnote{$https://github.com/deepinsight/insightface$} \cite{deng_arcface_2019},
Arcface-Focal\footnote{$https://github.com/ZhaoJ9014/face.evoLVe$} \cite{wang2021face}, and MagFace\footnote{$https://github.com/IrvingMeng/MagFace$} \cite{meng_magface_2021} matchers.


\section{Results and Discussion}
In this work, we evaluated all the possible mated pairs in Morph-II, AgeDB, and YFA datasets to construct a genuine distribution, while randomly selecting 1000 non-mated pairs for each sample to construct an imposter distribution in each dataset. In LAG dataset, we only evaluated the very challenging mated pairs between children and adults to represent the performance of the matchers in the children to adult large age-gap face verification task. The verification performance of CACD-VS dataset is evaluated using the positive and negative pairs presented in the dataset. Table II illustrates the TAR of the Facenet-V1, Facenet-V2, VGGFace, VGGFace2, ArcFace, ArcFace-Focal, and MagFace matchers at 0.01\% FAR. We can observe that MagFace, ArcFace, ArcFace-Focal, and Facenet-V1 matchers are outperforming the softmax based Facenet-V2, VGGFace and VGGFace2 matchers. This pattern is due to the discriminative nature (within-identity compactness) of these matchers vs the separable features observed in the softmax based matchers \cite{wen2019comprehensive}. As a result, we focus on the performance of these discriminative matchers for the rest of this work. 

We first evaluate each matcher on CACD\_VS, Morph-II, AgeDB, and LAG datasets to paint a picture of the cross-age verification performance of the state-of-the-art FR matchers. Our tests demonstrate a high TAR for all four matchers in the controlled and short age-gap adult dataset (MORPH-II). The best performing matcher (ArcFace-Focal) achieves 98.9\% TAR at 0.01\% FAR in this dataset. However, extending the verification task beyond short age-gap and controlled acquisition environment (CACD-VS and AgeDB), results in a decrease in the performance of all four matchers. By extending the maximum age-gap from 5 (Morph-II) to 10 years (CACD\_VS), we observe that while ArcFace-Focal and Facenet-V1 respectively maintain high TARs of 96.4\% and 94.4\%, ArcFace and MagFace respectively suffer from 19.8\% and 13.6\% drop in their TAR at 0.01\% FAR.

Further increasing the age-gaps up to the maximum of 90 years (AgeDB), results in 46.0\%, 58.9\%, 30.8\%, and 56.6\% drop in the respective TARs of Facenet-V1, ArcFace, ArcFace-Focal and MagFace at 0.01\% FAR. We observe that ArcFace-Focal shows more tolerance to larger age gaps in Agedb and considerably outperforms the next best matcher (Facenet-v1) by 19.7\% TAR at 0.01\% FAR. Finally, the LAG dataset allows us to extend the large age-gap verification task to the most challenging case of children to adult matching. Interestingly, in this case, we even observe a sharp drop (38.4\%) in the TAR of ArcFace-Focal in addition to all other matchers. Our results suggest that none of the evaluated matchers could perform well in children to adult large age-gap verification in-the-wild. 

Our result indicates a lower performance in the four top performing matchers in YFA (short age-gap, children) with respect to that of Morph-II (short age-gap, adults) confirming the challenging nature of the child face recognition. However, while ArcFace-Focal, ArcFace, and Facenet-V1 respectively show 13.5\%, 19.6\% and 26.2\% drop in the TAR at 0.01\% FAR, Magface only suffers from 3.5\% reduction in TAR and achieves 93.3\% TAR at 0.01\% FAR. We believe this higher verification performance is due to the quality-aware structure of the Magface within-identity distributions \cite{meng_magface_2021}. 

The Facenet-V1 matcher respectively achieves 67.9\% and 34.7\% TAR at 0.01\% FPR with 1 and 3 years of age-gap on CLF dataset \cite{deb2018longitudinal}. This matcher achieves 80.1\% and 43.4\% TAR at 0.01\% FPR at 1 and 3 years age-gap in YFA dataset. This increase in the performance might be due to the higher quality and controlled image acquisition environment of YFA or the difference between the ethnicity of subjects in YFA and CLF datasets. Furthermore, Table III illustrates the TAR as a function of age-gap for the top four best performing matchers in YFA dataset. Our result confirms the previously identified downward trend in the TAR of Facenet-V1 matcher \cite{deb2018longitudinal}. Additionally, Table III also reveals a noticeable degradation in the TAR even at age-gap of 6 months for Facenet-V1, ArcFace and ArcFace-Focal. For instance, the second-best performing matcher (ArcFace-Focal) shows 1.3\%, 3.6\%, 8.4\%, 1.0\%, and 5.7\% drops in TAR at 0.01\% FAR over the increasing age-gaps of 6 to 36 months. 
However, contrary to previous work, we observe that this reduction is not substantial in MagFace. For instance, we only observe a 0.9\% drop in the TAR at 0.1\% FAR over a 30 month age-gap, suggesting that accurate children face recognition may be practical using high-quality samples and quality-aware matchers for up to 3 years. 

\subsection{Match Score Modeling}
In this work, we utilize Linear Mixed Effect (LME) models to investigate the influence of factors affecting the Match Score (MS) of FR systems in children. LME models can be effectively applied to the multilevel analysis of longitudinal data \cite{lindstrom_newtonraphson_1988}. Such models can represent the variations observed in biometric systems as a combination of \textit{fixed} and \textit{random} effects, where the fixed effects represent the effects of explanatory variables on the inter-subject variations of the observed variable, while random effects quantify the intra-subject variations. Such characteristics make the LME models a viable candidate for modeling the changes in the MS of biometric systems using longitudinal data \cite{deb2018longitudinal, das_iris_2021}. We present and evaluate an LME model using the same explanatory variables proposed in \cite{deb2018longitudinal}. This model considers Enrolment Age (EA), age-gap between the enrolment and query ($\Delta{T}$), and Gender (G) as explanatory variables. Equation \ref{equ:model} presents our model to predict the MS of the children FR systems. Where,

\begin{equation}
\begin{split}
MS \sim \beta_{0} + \beta_{1}\Delta{T}+ \beta_{2}EA +\beta_{3}G + b_{0i}\\  
+ b_{1i}\Delta{T} + b_{2i}G
\label{equ:model}
\end{split}
\end{equation}

\begin{itemize}

    \item $\beta_{k}$ is the fixed regression coefficient for corresponding parameter, k.

    \item $b_{ki}$ is the random regression coefficient for corresponding parameter, k, for subject, i.
    
    \item $\beta_{0} + b_{0i}$ is the sum of fixed and subject specific random intercept corresponding to the initial state.
     
    \item $\beta_{k} + b_{ki}$ is the subject specific (i) gradient for the corresponding parameter k.
    
    \item $\Delta{T}$ denotes the age-gap between enrollment and query samples in years. $\Delta{T}$ has been considered for both fixed and random effects to account for possible variability in MS due to both intra-subjects and inter-subject effects of age-gap.
    
    \item EA denotes Enrolment Age (Years). EA has been considered only for fixed effect.
        
    \item G denotes the gender of the subject, encoded as 0 for Male and 1 for Female subjects. G is considered for both fixed and random effect.
    
\end{itemize}
Initially, we define two mutually exclusive subsets (\textit{Young} and \textit{Adult}) to independently analyze younger and older cohorts of AgeDB. Our \textit{Young} section includes mated pairs with enrolment age younger than 15 years old and query images up to 20 years old. We define the \textit{Adult} section as mated pairs with both enrolment and query images older than 20 years old. We conduct our analysis with the following hypothesis:

\begin{itemize}
    \item Null Hypothesis ($H_0$): There is no correlation between the individual predictors and the response variable (MS).
    \item Alternative Hypothesis ($H_A$): There is a correlation between the individual predictors and the response variable.
\end{itemize}
Tables IV shows the fixed effects from Morph-II, AgeDB, and YFA datasets. All statistical analysis in this work is carried out using the python \textit{statsmodels-0.12.0} library with Restricted Maximum Likelihood (REML) estimator \cite{seabold_statsmodels_2010}.

\subsubsection{Effect of Age-gap and Enrolment Age in Adult and Children}
We reject the null hypothesis with $p<0.001$ across matchers, confirming a significant decaying relationship between age-gap and MS in both adults and children. Our analysis with best performing matcher (MagFace), respectively indicates an estimated decrease in the MS, ($\beta_0$), due to age-gap, ($\beta_1$), of $-0.011 \pm 0.000$ (approx) and $-0.004 \pm 0.000$ (approx) with each year for adults in Morph-II and \textit{adult} subsection of AgeDB datasets. On the other hand, we observe a much higher estimated decrease in the MS, due to age-gap, ($\beta_1$) in \textit{Young} section of AgeDB ($-0.022 \pm 0.004$) and YFA dataset ($-0.033 \pm 0.003$). Our result indicates the consistency of this pattern across multiple matchers. However, we cannot observe a consistent and matcher independent pattern across the estimated effects of enrolment age ($\beta_2$) on the match scores of either adults or children.

\subsubsection{Effect of Gender}
We reject the null hypothesis with $p<0.001$ across all matchers in the Morph-II dataset (short age-gap, adult), confirming a significant decaying relationship between gender and MS in this dataset. Our model respectively indicates an estimated decrease in the MS, due to gender ($\beta_3$), of $-0.010 \pm 0.002$ (approx) $-0.038 \pm 0.002$ (approx), $-0.025 \pm 0.002$ (approx) for female subjects in Magface, ArcFace-Focal, and Facenet-V1 matchers. However, our result confirms the findings of \cite{deb2018longitudinal}, as we do not observe a significant relationship between gender and MS either in AgeDB or YFA.

\section{Limitations and Future Work}
Our result confirms the lower verification performance observed in previous work with in-the-wild datasets and older deep face matchers even with high-quality samples. Additionally, our analysis confirms a statistically significant and matcher independent decaying relationship between the evaluated matchers and the age-gap in children. However, contrary to previous work using in-the-wild data, our verification performance analysis (Table-II) indicates that the combination of high-quality samples and state-of-the-art quality-aware deep face matchers could be a viable solution for children face recognition up to 3 years. Our comparison between the verification performance of ArcFace, ArcFace-Focal and MagFace suggests that the higher performance observed in the latter two matchers might be due to the adaptive hard sample mining process utilized in both Focal-loss and MagFace. Future work can further investigate the fusion of the best performing matchers. Additionally, the within-identity structure of these matchers can be utilized for linear and non-linear feature projection methods to possibly even further increase the performance.
\bibliographystyle{IEEEtran}
\bibliography{references_face}

\begin{thebibliography}{10}
\providecommand{\url}[1]{#1}
\csname url@samestyle\endcsname
\providecommand{\newblock}{\relax}
\providecommand{\bibinfo}[2]{#2}
\providecommand{\BIBentrySTDinterwordspacing}{\spaceskip=0pt\relax}
\providecommand{\BIBentryALTinterwordstretchfactor}{4}
\providecommand{\BIBentryALTinterwordspacing}{\spaceskip=\fontdimen2\font plus
\BIBentryALTinterwordstretchfactor\fontdimen3\font minus
  \fontdimen4\font\relax}
\providecommand{\BIBforeignlanguage}[2]{{%
\expandafter\ifx\csname l@#1\endcsname\relax
\typeout{** WARNING: IEEEtran.bst: No hyphenation pattern has been}%
\typeout{** loaded for the language `#1'. Using the pattern for}%
\typeout{** the default language instead.}%
\else
\language=\csname l@#1\endcsname
\fi
#2}}
\providecommand{\BIBdecl}{\relax}
\BIBdecl

\bibitem{keahey_lessons_2020}
K.~Keahey, J.~Anderson, Z.~Zhen, P.~Riteau, P.~Ruth, D.~Stanzione, M.~Cevik,
  J.~Colleran, H.~S. Gunawi, and C.~Hammock, ``Lessons learned from the
  chameleon testbed,'' in \emph{Proceedings of the 2020 {USENIX} {Annual}
  {Technical} {Conference} ({USENIX} {ATC} '20)}.\hskip 1em plus 0.5em minus
  0.4em\relax USENIX Association, 2020, pp. 219--233.

\bibitem{lanitis_survey_2010}
A.~Lanitis, ``A survey of the effects of aging on biometric identity
  verification,'' \emph{International Journal of Biometrics}, vol.~2, no.~1,
  p.~34, 2010.

\bibitem{otto_how_2012}
C.~Otto, H.~Han, and A.~Jain, ``How does aging affect facial components?'' in
  \emph{European {Conference} on {Computer} {Vision}}.\hskip 1em plus 0.5em
  minus 0.4em\relax Springer, 2012, pp. 189--198.

\bibitem{ricanek_morph_2006}
K.~Ricanek and T.~Tesafaye, ``Morph: {A} longitudinal image database of normal
  adult age-progression,'' in \emph{7th {International} {Conference} on
  {Automatic} {Face} and {Gesture} {Recognition} ({FGR06})}.\hskip 1em plus
  0.5em minus 0.4em\relax IEEE, 2006, pp. 341--345.

\bibitem{huang_labeled_2014}
G.~B. Huang and E.~Learned-Miller, ``Labeled faces in the wild: {Updates} and
  new reporting procedures,'' \emph{Dept. Comput. Sci., Univ. Massachusetts
  Amherst, Amherst, MA, USA, Tech. Rep}, vol.~14, no. 003, 2014.

\bibitem{grother_face_2019}
\BIBentryALTinterwordspacing
P.~Grother, M.~Ngan, and K.~Hanaoka, ``\BIBforeignlanguage{en}{Face
  {Recognition} {Vendor} {Test} ({FRVT}) part 2 :: identification},'' National
  Institute of Standards and Technology, Gaithersburg, MD, Tech. Rep. NIST IR
  8271, Sep. 2019. [Online]. Available:
  \url{https://nvlpubs.nist.gov/nistpubs/ir/2019/NIST.IR.8271.pdf}
\BIBentrySTDinterwordspacing

\bibitem{zheng_cross-age_2017}
T.~Zheng, W.~Deng, and J.~Hu, ``Cross-age lfw: {A} database for studying
  cross-age face recognition in unconstrained environments,'' \emph{arXiv
  preprint arXiv:1708.08197}, 2017.

\bibitem{chen_face_2015}
B.-C. Chen, C.-S. Chen, and W.~H. Hsu, ``Face recognition and retrieval using
  cross-age reference coding with cross-age celebrity dataset,'' \emph{IEEE
  Transactions on Multimedia}, vol.~17, no.~6, pp. 804--815, 2015, publisher:
  IEEE.

\bibitem{moschoglou_agedb_2017}
S.~Moschoglou, A.~Papaioannou, C.~Sagonas, J.~Deng, I.~Kotsia, and
  S.~Zafeiriou, ``Agedb: the first manually collected, in-the-wild age
  database,'' in \emph{Proceedings of the {IEEE} {Conference} on {Computer}
  {Vision} and {Pattern} {Recognition} {Workshops}}, 2017, pp. 51--59.

\bibitem{bianco_large_2017}
S.~Bianco, ``Large age-gap face verification by feature injection in deep
  networks,'' \emph{Pattern Recognition Letters}, vol.~90, pp. 36--42, 2017,
  publisher: Elsevier.

\bibitem{ricanek_jr_craniofacial_2008}
K.~Ricanek~Jr, A.~Sethuram, E.~K. Patterson, A.~M. Albert, and E.~J. Boone,
  ``Craniofacial aging,'' \emph{Wiley Handbook of Science and Technology for
  Homeland Security}, pp. 1--27, 2008, publisher: Wiley Online Library.

\bibitem{ramanathan_modeling_2006}
N.~Ramanathan and R.~Chellappa, ``Modeling age progression in young faces,'' in
  \emph{2006 {IEEE} {Computer} {Society} {Conference} on {Computer} {Vision}
  and {Pattern} {Recognition} ({CVPR}'06)}, vol.~1.\hskip 1em plus 0.5em minus
  0.4em\relax IEEE, 2006, pp. 387--394.

\bibitem{ricanek_review_2015}
K.~Ricanek, S.~Bhardwaj, and M.~Sodomsky, ``A review of face recognition
  against longitudinal child faces,'' \emph{BIOSIG 2015}, 2015, publisher:
  Gesellschaft für Informatik eV.

\bibitem{srinivas_face_2019}
N.~Srinivas, K.~Ricanek, D.~Michalski, D.~S. Bolme, and M.~King, ``Face
  recognition algorithm bias: {Performance} differences on images of children
  and adults,'' in \emph{Proceedings of the {IEEE}/{CVF} {Conference} on
  {Computer} {Vision} and {Pattern} {Recognition} {Workshops}}, 2019, pp. 0--0.

\bibitem{deb2018longitudinal}
D.~Deb, N.~Nain, and A.~K. Jain, ``Longitudinal study of child face
  recognition,'' in \emph{2018 International Conference on Biometrics
  (ICB)}.\hskip 1em plus 0.5em minus 0.4em\relax IEEE, 2018, pp. 225--232.

\bibitem{schroff_facenet:_2015}
F.~Schroff, D.~Kalenichenko, and J.~Philbin, ``Facenet: {A} unified embedding
  for face recognition and clustering,'' in \emph{Proceedings of the {IEEE}
  conference on computer vision and pattern recognition}, 2015, pp. 815--823.

\bibitem{guo2016ms}
Y.~Guo, L.~Zhang, Y.~Hu, X.~He, and J.~Gao, ``Ms-celeb-1m: A dataset and
  benchmark for large-scale face recognition,'' in \emph{European conference on
  computer vision}.\hskip 1em plus 0.5em minus 0.4em\relax Springer, 2016, pp.
  87--102.

\bibitem{michalski_impact_2018}
D.~Michalski, S.~Y. Yiu, and C.~Malec, ``The impact of age and threshold
  variation on facial recognition algorithm performance using images of
  children,'' in \emph{2018 {International} {Conference} on {Biometrics}
  ({ICB})}.\hskip 1em plus 0.5em minus 0.4em\relax IEEE, 2018, pp. 217--224.

\bibitem{deng_arcface_2019}
J.~Deng, J.~Guo, N.~Xue, and S.~Zafeiriou, ``Arcface: {Additive} angular margin
  loss for deep face recognition,'' in \emph{Proceedings of the {IEEE}/{CVF}
  {Conference} on {Computer} {Vision} and {Pattern} {Recognition}}, 2019, pp.
  4690--4699.

\bibitem{wang2021face}
Q.~Wang, P.~Zhang, H.~Xiong, and J.~Zhao, ``Face. evolve: A high-performance
  face recognition library,'' \emph{arXiv preprint arXiv:2107.08621}, 2021.

\bibitem{meng_magface_2021}
Q.~Meng, S.~Zhao, Z.~Huang, and F.~Zhou, ``\BIBforeignlanguage{en}{{MagFace}:
  {A} {Universal} {Representation} for {Face} {Recognition} and {Quality}
  {Assessment}},'' 2021, pp. 14\,225--14\,234.

\bibitem{parkhi_deep_2015}
O.~M. Parkhi, A.~Vedaldi, and A.~Zisserman, ``Deep face recognition,'' 2015,
  publisher: British Machine Vision Association.

\bibitem{cao2018vggface2}
Q.~Cao, L.~Shen, W.~Xie, O.~M. Parkhi, and A.~Zisserman, ``Vggface2: A dataset
  for recognising faces across pose and age,'' in \emph{2018 13th IEEE
  international conference on automatic face \& gesture recognition (FG
  2018)}.\hskip 1em plus 0.5em minus 0.4em\relax IEEE, 2018, pp. 67--74.

\bibitem{anjos_beat_2017}
A.~Anjos, L.~El-Shafey, and S.~Marcel, ``{BEAT}: {An} open-source web-based
  open-science platform,'' \emph{arXiv preprint arXiv:1704.02319}, 2017.

\bibitem{lindstrom_newtonraphson_1988}
M.~J. Lindstrom and D.~M. Bates, ``Newton—{Raphson} and {EM} algorithms for
  linear mixed-effects models for repeated-measures data,'' \emph{Journal of
  the American Statistical Association}, vol.~83, no. 404, pp. 1014--1022,
  1988, publisher: Taylor \& Francis.

\bibitem{zhang_joint_2016}
K.~Zhang, Z.~Zhang, Z.~Li, and Y.~Qiao, ``Joint face detection and alignment
  using multitask cascaded convolutional networks,'' \emph{IEEE Signal
  Processing Letters}, vol.~23, no.~10, pp. 1499--1503, 2016, publisher: IEEE.

\bibitem{wen2019comprehensive}
Y.~Wen, K.~Zhang, Z.~Li, and Y.~Qiao, ``A comprehensive study on center loss
  for deep face recognition,'' \emph{International Journal of Computer Vision},
  vol. 127, no.~6, pp. 668--683, 2019.

\bibitem{das_iris_2021}
P.~Das, L.~Holsopple, D.~Rissacher, M.~Schuckers, and S.~Schuckers, ``Iris
  {Recognition} {Performance} in {Children}: {A} {Longitudinal} {Study},''
  \emph{IEEE Transactions on Biometrics, Behavior, and Identity Science},
  vol.~3, no.~1, pp. 138--151, 2021, publisher: IEEE.

\bibitem{seabold_statsmodels_2010}
S.~Seabold and J.~Perktold, ``Statsmodels: {Econometric} and statistical
  modeling with python,'' in \emph{Proceedings of the 9th {Python} in {Science}
  {Conference}}, vol.~57.\hskip 1em plus 0.5em minus 0.4em\relax Austin, TX,
  2010, p.~61.

\end{thebibliography}
\end{document}